\documentclass[conference]{IEEEtran}
\ifCLASSINFOpdf
 \usepackage{graphics} 
	\usepackage{epsfig} 
	\usepackage{mathptmx} 
	\usepackage{times} 
	\usepackage{amsmath} 
	\usepackage{amssymb}  
	\usepackage{textcomp}
	\usepackage{multirow}
	\usepackage{colortbl}
	\usepackage{xcolor}
	\usepackage{hhline}
	\usepackage{subfigure}
	\graphicspath{{}}
\else
\fi

\usepackage{algorithm}
\usepackage{algpseudocode}
\usepackage{url}
\usepackage{hyperref}

\begin{document}
%
\title{Deep Grid Net (DGN): A Deep Learning System for Real-Time Driving Context Understanding}



%
\author{Liviu A. Marina,
Bogdan Trasnea,
Cocias Tiberiu, 
Andrei Vasilcoi,
Florin Moldoveanu and
Sorin M. Grigorescu \\
Elektrobit Automotive Romania\\
 Transilvania University of Brasov, Department of Automation,
Brasov, 500036 Romania\\ Email: marina.liviu.alexandru@unitbv.ro}

\maketitle

\begin{abstract}
Grid maps obtained from fused sensory information are nowadays among the most popular approaches for motion planning for autonomous driving cars. In this paper, we introduce \textit{Deep Grid Net} (DGN), a deep learning (DL) system designed for understanding the context in which an autonomous car is driving. DGN incorporates a learned driving environment representation based on Occupancy Grids (OG) obtained from raw Lidar data and constructed on top of the Dempster-Shafer (DS) theory. The predicted driving context is further used for switching between different driving strategies implemented within EB robinos, Elektrobit’s Autonomous Driving (AD) software platform. Based on genetic algorithms (GAs), we also propose a neuroevolutionary approach for learning the tuning hyperparameters of DGN. The performance of the proposed deep network has been evaluated against similar competing driving context estimation classifiers.
\end{abstract}


\IEEEpeerreviewmaketitle

\section{Introduction}
\label{sec:introduction}

In order for a \textit{Highly Autonomous Driving} (HAD) system to select optimal driving strategies, it must first understand the context in which the vehicle is driving. For example, an HAD system deploys different driving strategies when the ego-car is driving on the highway, as opposed to driving in the inner-city.

\begin{figure}
	\centering
	\begin{center}
		\includegraphics[scale=0.64]{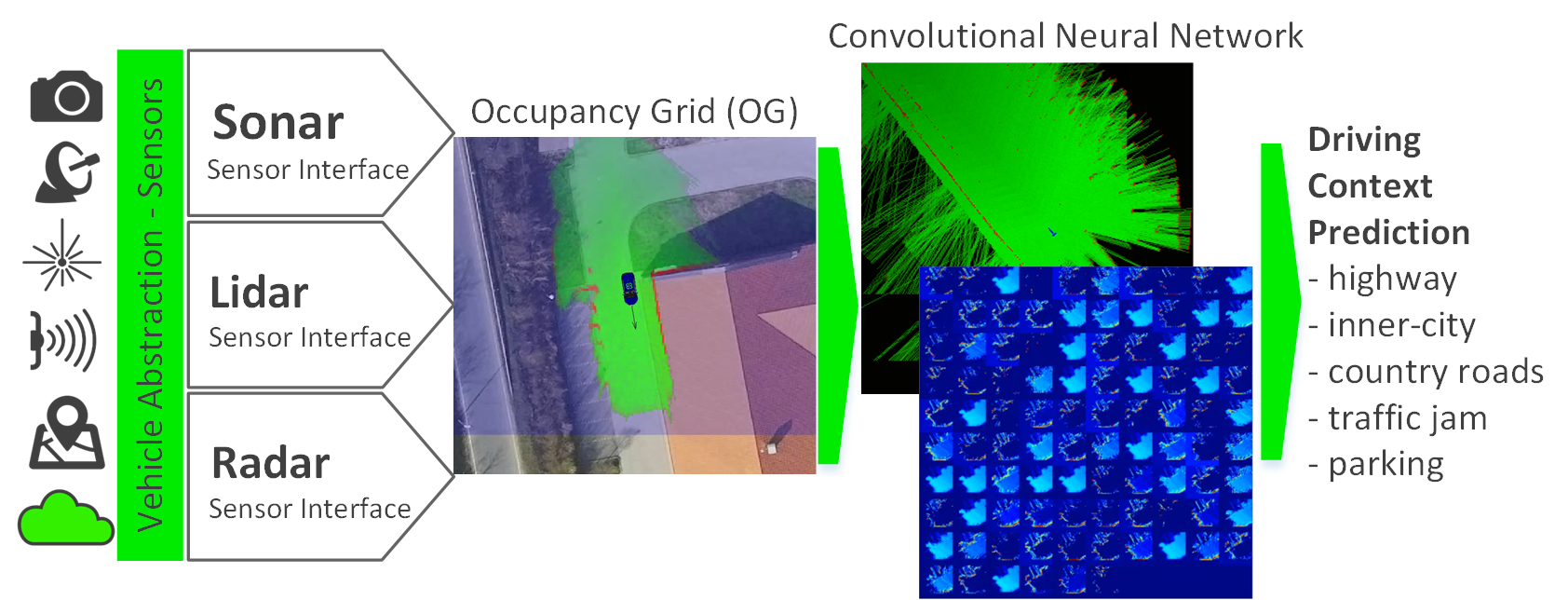}
		\caption{DGN's workflow diagram. Lidar sensory data streams are converted into OGs, which are furthered parsed by a \textit{Convolutional Neural Network} (CNN). The final layer provides the driving context as a five classes probabilistic output (highway, inner-city, country roads, traffic jam situations and parking lots). A video presenting DGN's functionality can be found at this \href{https://www.youtube.com/watch?v=vDafgRWE_Jo&feature=youtu.be}{link} }.
        \label{fig:DGN_Architecture}
	\end{center}
\end{figure}

Illustrated in Fig.~\ref{fig:DGN_Architecture}, the \textit{Deep Grid Net} (DGN) algorithm predicts the driving context by analyzing local OGs. The main advantage in using OGs, as opposed to image-based context understanding, is that the search space is highly reduced since the information represented in OGs is much lower than in the case of images. 
In this work, the occupancy grids are built from data acquired from Lidar sensors, mounted on the front and rear sections of the ego-car.

The DGN algorithm is deployed within Elektrobit's HAD software framework, coined EB robinos, a functional software architecture that manages the complexity of autonomous driving. DGN offers a robust real-time estimation of the driving context mapped to five classes: driving on the highway, driving in the inner-city, driving on the country roads, driving in ares with traffic jam situations and parking.

Deep neural networks (DNN) were chosen to encode the traffic scene due to their generalization capabilities. The number of configuration parameters of a DNN, also known as hyperparameters, increased together with the size and complexity of the networks. In order to overcome the manually tuning of these hyperparameters, we build on top of the authors previous work on one-shot learning using neuroevolutionary algorithms~\cite{SorinGr} and propose an approach for the automatic computation of hyperparameters during training. 

The main contributions of this paper can be summarized as:

\begin{itemize}
    \item Introduction of the DGN architecture, encoding a learned grid-based representation of the traffic scene;
    \item DGN's hyperparameters tuning using GAs;
    \item Deployment of DGN into the EB robinos autonomous driving software stack.
\end{itemize}

The rest of the paper is organized as follows: an overview of related work is given in Section \ref{sec:background_and_motivation}, while the \textit{DGN} system is presented in Section~\ref{sec:method}. A description of the training strategy, and the evaluation of DGN's performance are given in Section~\ref{sec:exp_res}. Finally, the conclusions are stated in Section~\ref{sec:conclusion}.

\section{Background and Motivation}
\label{sec:background_and_motivation}

Occupancy grids are widely used to map indoor spaces in autonomous navigation for self-driving agents. In \cite{SemanticLabels}, \textit{Convolutional Neural Networks} (CNNs) have been trained on 2D range sensory data for the semantic labeling of places in unseen environments. Whithin this approach OGs created from Lidar scans are converted to gray images and used classify between three classes, that is, room, corridor, and doorway. 

Several papers reported OGs constructed from the interaction of a robot with its surrounding environment. \textit{Recurrent Neural Networks} (RNN) have been used by Ondruska \cite{EndtoEndOdruska} for tracking and classifying the surroundings of a robot placed in a dynamic and partially observable environment. A RNN filters the input stream composed of raw laser measurements in order to infer the objects locations together with their identity. The algorithm in \cite{EndtoEndOdruska} takes inspiration from Deep Tracking \cite{DeepTrackingOdruska}.

In \cite{DynamicOccupancy}, an environment modeled with a Bayesian filtering technique is processed through a DNN, with the purpose of obtaining a long-term driving situation prediction for intelligent vehicles. This work is based on the principles stated by Nuss in \cite{RandomNuss, FusionNuss}, where raw Radar and Laser data is parsed through a fusion layer. The algorithm predicts future static and dynamic objects using a CNN trained on occupancy grids.

Although OGs are common tools in robotics, there are a few cases where DL techniques are used for real-time environment perception using OGs. 
In \cite{EnvironmentDezert}, an improved version of the \textit{Proportional Conflict Redistribution} rule '\#'6 (PCR6), taking into account Zhang’s degree of intersection of focal elements \cite{PCRSmarandache}, was used on Lidar data.

In \cite{Seeger}, OGs and DNNs have been applied to outdoor driving scene classification. A major differences with respect to our work is that the classifier in \cite{Seeger} estimates only four driving classes based on OGs which are constructed by accumulating data over time. In our work, we compute OGs in real-time, during the movement of the vehicle, in order to classify five road types.

\section{Methodology}
\label{sec:method}

\begin{figure*}
	\centering
	\begin{center}
		\includegraphics[scale=1.52]{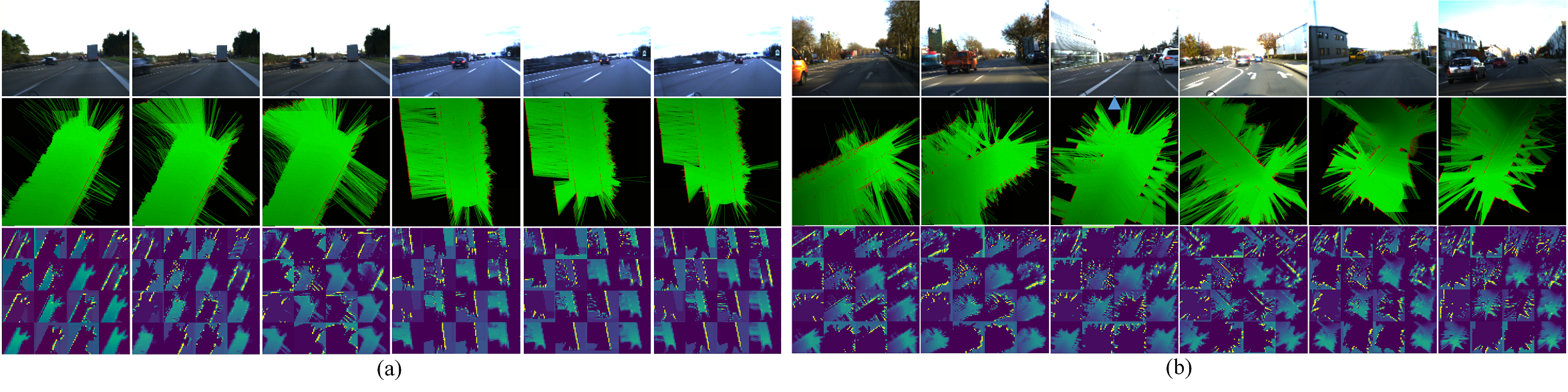}
		\caption{Examples of real-world OGs samples for highways (a) and inner-city streets (b). The top images in each group represent snapshots of the driving environment, together with their respective OG and the activations of the first CNN layer.}
        \label{fig:grid_examples}
	\end{center}
\end{figure*}


\subsection{Problem space}
\label{sec:problem_space}
                                                                                                                                                                             
The DGN algorithm is mainly composed of three elements: (i) an OG fusion algorithm, (ii) a DNN used for parsing the OG in real-time and (iii) an evolutionary algorithm used for selecting the optimal DNN hyperparameters set. The outcome obtained from DGN is a driving context classification, mapped to five classes: inner city (IC), country road (CR), parking lot (PL), highway (HW) and traffic jam (TJ).

The OG training dataset $D$ is used to calculate the optimal DGN hypothesis $h_{DGN}$, which encodes the deep network's structure and weights. We define our problem space within the following Bayesian framework:

\begin{equation}
	P(h|D)=\frac{P(D|h)P(h)}{P(D)}
\end{equation}

\noindent where $P(h)$ is the prior probability over $h$,  $P(D) = \int_{h}P(D|h)P(h)$ is the training data probability, and $P(h|D)$ is the likelihood of $h$ given $D$. $P(D|h)$ is the data likelihood over a given hypothesis. The maximum a posteriori (MAP) hypothesis $h_{MAP}$,  using Bayes theorem, can be defined as:

\begin{equation}
	h_{MAP}=\operatorname*{argmax}_{h \in \mathcal{H}}P(D|h)P(h).
\end{equation}

Assuming that all hypotheses are equally probable, we can choose a \textit{Maximum Likelihood} (ML) approach for training:

\begin{equation}
	h_{ML}=\operatorname*{argmax}_{h \in \mathcal{H}}P(D|h) = \operatorname*{argmax}_{h \in \mathcal{H}}L(h)
\end{equation}

The training samples are considered to be independently identically distributed, thus satisfying the following statement:

\begin{equation}
\label{eq_ML}
	P(D|h) =\prod\limits_{i=1}^{m}P(\langle x_i, y_i \rangle | h)=\prod\limits_{i=1}^{m}P(y_i | x_i;h)P(x_i)
\end{equation}

Maximizing the Eq.~\ref{eq_ML} is equivalent with the maximization of the logarithmic function $\log L(h)$, where the term $\sum\limits_{i=1}^{m}\log P(x_i)$ depends on $D$, but not on $h$ and it can be ignored:

\begin{equation}
	\log L(h_{DGN}) =\sum\limits_{i=1}^{m}\log P(y_i | x_i;h)
	\label{eq:mle}
\end{equation}

\subsection{Occupancy Grids}
\label{sec:occupancy_grids}

Occupancy Grids are the data source for calculating the optimal DGN hypothesis $h_{DGN}$. In our work, the grids used for driving context classification where built using the \textit{Dempster-Shafer} (DS) theory \cite{Shafer}.

From the different fusion rules proposed in literature \cite{AdvancesSmarandache}, the DS rule was most suited for our work. The issue which arises here is how to combine two independent sets of probability mass assignments with specific situations. The joint mass is calculated from the $m_1(.)$ and $m_2 (.)$ sets of masses. The DS combination is defined by taking $m_{1,2}^{DS} (\Phi)=0$ for all $X \neq \Phi$:

\begin{equation}
	m_{1,2}^{DS} \triangleq \frac{1}{1-m_{1,2}(\phi)} \sum\limits_{\substack{X_1,X_2 \in 2^\Omega\\X_1 \cap X_2 \neq \phi}}\prod\limits_{i=1}^2m_i(X_i)
\end{equation}

\noindent where $m_{1,2} (\phi)$ measures the amount of conflict between the two mass sets, and $1-m_{1,2} (\phi)$ is a normalization constant. 
 
The idea behind OGs is the environment’s division into 2D cells, each cell representing the probability, of occupation. Each cell is color-coded, green pixels representing the free space, red marking the occupied cells (or obstacles) and black models the unknown occupancy. The color intensity represents the degree of occupancy. The grid content is updated over and over again, in real-time, with each sensory measurement. Examples of labelled OGs are shown in Fig.~\ref{fig:grid_examples}.

\subsection{Neuroevolutionary Training of DGN}
\label{sec:genetic_algorithms}

\begin{algorithm}
	\caption{DGN's training procedure in pseudocode.}\label{alg:neuroevolve}
	\begin{algorithmic}[1]
		\Procedure{Train}{$\mathcal{P}$}
		\For{\texttt{($\theta$,$h_{DGN}$,$t$) $\in$ $\mathcal{P}$ }}
				\While{\texttt{not end of training}}
			   \State $\theta \gets backprop (\theta | h_{DGN})$
			   \State $h_{DGN} \gets {eval (h_{DGN})}$ 
			   \State $h_{DGN} \gets {explore (h_{DGN}, \mathcal{P} )}$
			   \State \texttt {update $\mathcal{P}$ with ($h_{DGN}$, $\theta$, $t+1$)}
				\EndWhile
		\EndFor
		\State \textbf{return} \texttt{top 5 $h_{DGN}$ individuals in  $\mathcal{P}$}   
		\EndProcedure
	\end{algorithmic}
	\label{alg:alg1}
\end{algorithm}

\textit{Genetic Algorithms} (GAs) \cite{Eiben_GA} are a metaheuristic optimization method, belonging to a broader class of evolutionary algorithms. The evolutionary training process starts from an initial set of solutions, or population $\mathcal{P}$, where every solution is given by a set of properties, called genes. A solution is also called an individual $h_{DGN} \in \mathcal{P}$. 

GAs are used in our work for finding the optimal hyperparameters set encoding $h_{DGN}^*$, that is, the optimal number of neurons in each layer, most suitable optimizer and the best cost function for backpropagation. This allows us to determine the smallest DNN structure, which can deliver accurate results, as well as real-time processing capabilities \cite{popbased_NN}. An $eval$ function has been defined to find the optimal set of parameters:

\begin{equation}
	\theta^* =\operatorname*{argmax}_{\theta \in \Theta}  eval(\theta). 
\end{equation}

The proposed training method optimizes over a hyperparameters solutions space, aiming at calculating the top individuals $h_{DGN} \in \mathcal{P}$ based on their fitness value:

\begin{equation}
\label{eq_hyperparam}
	h_{DGN}^* = \operatorname*{argmax}_{h_{DGN} \in \mathcal{P}} eval ( backprop (\theta | h_{DGN}) )
\end{equation}

The optimal structure of the network is evaluated within the training loop for a given set of weights $\theta$, DGN individual $h_{DGN}$ and training step $t$. The weights $\theta$ are calculated using classical backpropagation, according to the maximum likelihood estimation defined in Eq.~\ref{eq:mle}:

\begin{equation}
	\theta \gets {backprop  (\theta | h_{DGN})}. 
	\label{eq:backprop}
\end{equation}

Once the training in Eq.~\ref{eq:backprop} is completed, the hyperparameters are evaluated based on $h_{DGN}$ using the $eval(\cdot)$ fitness function. The new set of hyperparameters are calculated by exploring the solution space with the help of the $explore (h_{DGN}, \mathcal{P} )$ procedure. The training loop stops after 15 training epochs and returns the top 5 individuals, which have the highest fitness value, approach presented within the Algorithm~\ref{alg:alg1} pseudocode.

\subsection{DGN Architecture}
\label{sec:dgn_architectures}

The OGs computed with the above-described method represents the input to a CNN, which constructs a grid representation of the driving environment. The neural network topology is written in Keras, on top of the TensorFlow library \cite{chollet2015keras}. 

The DGN architecture has been developed for deployment within EB robinos, where smaller activation maps are required in order to achieve real-time performance. The DGN's topology consists of two convolutional layers with $32$ and $64$ kernel filters, respectively. The convolutional kernel has been reduced to a $9\times 9$, respectively $5 \times 5$ size for the first two network's layers. \textit{Rectified Linear Unit} (ReLu) filters each convolution, followed by a normalization layer and a pooling operation. The network also contains three fully connected (FC) layers linked to a final Softmax activation function which calculates the driving context probabilities. In order to reduce the model overfitting, Dropout layers were added.

\section{Experimental results}
\label{sec:exp_res}

\begin{figure}
	\centering
	\begin{center}
		\includegraphics[scale=0.560]{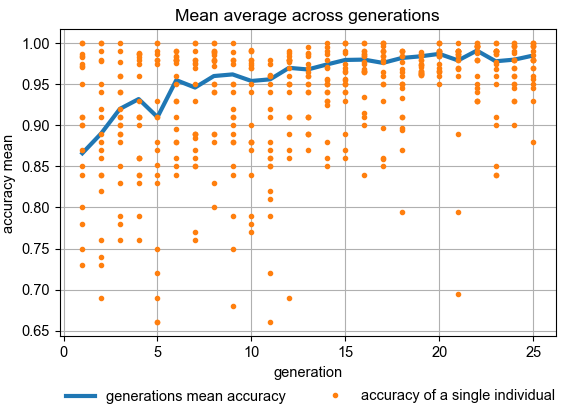}
		\caption{Fitness function evolution during training.}
        \label{fig:GA_acc}
	\end{center}
\end{figure}

\subsection{Training strategy}
\label{sec:training_strategy}

The data was collected on several road types in Germany, using a test vehicle equipped with a front camera (Continental MFC430), a front and rear lidar (Quanergy M8). 

The sensory data streams are fused into OGs having a size of $125 \times 125$ and a resolution of $0.25m$ per cell. The data samples are saved during driving at time intervals ranging between 50 \textit{ms} and 90 \textit{ms} per cycle. Approximately 60.000 samples were obtained, containing different scenarios types: country roads, highways, inner city, parking lots, or traffic jam situations.  From the total amount of samples, 65\%  were used for training, 20\% for validation and 15\% for testing.

The classification model was trained from scratch, using a learning rate $\alpha$ of 0.0001 for the backpropagation algorithm.


Our NN structure was determined based on the algorithm described in the Section~\ref{sec:genetic_algorithms}. The hyperparameters set used during training consists in \textit{optimizers} (rmsprop, adam, SGD, adagrad, adadelta, adamax, nadam),  \textit{loss functions} (categorical crossentropy, mean squared error) and \textit{number of neurons} (16, 32, 64, 128, 258)

The fitness function evolution can be seen in Fig.~\ref{fig:GA_acc}. The GA evolved the DGN's hyperparameters with a generation restraint value of $10$, each generation consisting of $20$ individuals. An individual represents a DGN neural network, with the hyperparameters selected by the neuroevolutionary algorithm.

An average classification accuracy is measured after each generation. When the last generation is reached, the individual with the best score is selected as $h_{DGN}^*$. With our training method, we have reached a value of $98.5\%$ fitness accuracy. The top network structure contains 64 and 32 neurons for the first and second FC layers, respectively, and uses \textit{categorical crossentropy} as loss function and \textit{adam} as optimizer.

\subsection{Accuracy Evaluation}
\label{sec:acc_evaluation}

The classification performance is summarized in the confusion matrix from Table~\ref{table_confMatrix}, where slight differences in the per-class performance are visible. 
The classes \textit{traffic jam} and \textit{parking lot} present a higher detection accuracy since its respective occupancy grids have a more distinctive structure.

\begin{table}[h]
	\caption{Confusion matrix evaluation}
	\label{table_confMatrix}
	\begin{center}
		\renewcommand{\arraystretch}{1.5}
		\begin{tabular}{cc|c|c|c|c|c|}
			\cline{3-7}
			& & \multicolumn{5}{c|}{\textbf{Actual class}} \\
			\cline{3-7}
			& &IC& CR & HW &PL &TJ \\ \hhline{-|-|-|-|-|-|-|}
			\multicolumn{1}{ |c }{\multirow{5}{*}{\rotatebox[origin=c]{90}{\parbox[c]{2cm}{\centering \textbf{Predicted Class}}}} } &
			\multicolumn{1}{ |c| }{IC} & 0.977 \cellcolor{green!95} &0.0201 & 0.0026 & 0.0001 & 0.0001 \\ \hhline{|~|-|-|-|-|-|-|}
			\multicolumn{1}{ |c }{} &
			\multicolumn{1}{ |c| }{CR} & 0.0027 & 0.992 \cellcolor{green!95} & 0.0051 & 0.0008 & 0.0003 \\ \hhline{|~|-|-|-|-|-|-|}
			\multicolumn{1}{ |c }{} &
			\multicolumn{1}{ |c| }{HW} & 0.0061 & 0.0259 & 0.967 \cellcolor{green!95} & 0.0004 & 0.0002 \\ \hhline{|~|-|-|-|-|-|-|}
			\multicolumn{1}{ |c }{} &
			\multicolumn{1}{ |c| }{PL} & 0.0001 & 0 .0004 & 0.0001 & 0.997 \cellcolor{green!95}& 0.0002 \\ \hhline{|~|-|-|-|-|-|-|}
			\multicolumn{1}{ |c }{} &
			\multicolumn{1}{ |c| }{TJ} & 0.006 & 0.003 & 0.0001 & 0.002 & 0.99 \cellcolor{green!95} \\ \cline{1-7}
		\end{tabular}
	\end{center}
\end{table}

A comparison of DGN's accuracy against state-of-the-art methods is presented in Table~\ref{table_conmAcc}. The competitors are several network topologies, like AlexNet \cite{AlexNet}, or GoogleLeNet \cite{GoogleLeNet}, as well as the algorithm from~\cite{Seeger}. All algorithms were tested with respect to the same testing data. The classification results obtained with DGN are clearly higher than the ones delivered by the competing estimators.

\begin{table}
	\caption{Comparison of driving context classification accuracy}
	\label{table_conmAcc}
	\begin{center}
		\renewcommand{\arraystretch}{1}
		\begin{tabular}{c|c|c|c|c}
			\textbf{Method} & \textbf{Accuracy} & \textbf{Recall} & \textbf{Precision} & \textbf{F-measure}\\
			\hline
			LeNET  & 0.88 & 0.91 & 0.93 & 0.92\\
			\hline
			GoogLeNet  & 0.94 & 0.96 & 0.97 & 0.97\\
			\hline
			ResNet  & 0.9 & 0.92 & 0.94 & 0.928 \\
			\hline
			AlexNet  & 0.95 & 0.95 & 0.96 & 0.96 \\
			\hline
			Seeger & 0.89 & 0.9 & 0.92 & 0.92 \\
			\hline
			\textbf{DGN} & \textbf{0.96} &  \textbf{0,985} & \textbf{0,984} & \textbf{0,984}\\
		\end{tabular}
	\end{center}
\end{table} 

Apart from its high classification accuracy, one other advantage of DGN is represented by the detection speed of the algorithm, making it suitable for real-time applications, like the EB robinos HAD platform. DGN runs on single OG sample, without the need to accumulate grid data over time, as required by the method in \cite{Seeger}.

\section{Conclusion}
\label{sec:conclusion}

In this paper, we have introduced \textit{DGN}, which is a solution for driving context understanding, required by behavior arbitration components within HAD systems. It has been designed to infer the driving context directly from OGs, as opposed to traditional image based methods. We were able to show that a simplified CNN topology is sufficient to classify in real-time between different types of OGs, without the need of training large networks, such as AlexNet, or GoogLeNet.


\ifCLASSOPTIONcaptionsoff
  \newpage
\fi

\bibliographystyle{IEEEtran}
\bibliography{references}

\end{document}